\documentclass[english]{article} 
\usepackage{latexsym,amsmath,url,caption2,epsfig}
\usepackage{amsfonts,euscript}
\usepackage{psfig}

\usepackage{hyperref}
\usepackage[latin9]{inputenc}
\usepackage{amsmath}
\usepackage{amssymb}
\usepackage[square,numbers]{natbib}
\usepackage{graphicx}

\usepackage[T1]{fontenc}
\usepackage[latin9]{inputenc}
\usepackage{array}
\usepackage{multirow}
\usepackage{amsthm}
\usepackage{amsmath}
\usepackage{amssymb}
\usepackage{algorithm}
\usepackage{algorithmic}

\makeatletter
\theoremstyle{plain}
\newtheorem{thm}{\protect\theoremname}

\newcommand\independent{\protect\mathpalette{\protect\independenT}{\perp}}
\def\independenT#1#2{\mathrel{\rlap{$#1#2$}\mkern2mu{#1#2}}}

\makeatother

\usepackage{babel}
\providecommand{\theoremname}{Theorem}

\title{Sparse Quadratic Discriminant Analysis \\
and Community Bayes}

\date{}

\author{
Ya Le \\
Department of Statistics\\
Stanford University\\
\texttt{yle@stanford.edu} \\
\and
Trevor Hastie \\
Department of Statistics \\
Stanford University \\
\texttt{hastie@stanford.edu} \\
}

\makeatother

\begin{document}
\maketitle

\begin{abstract}
We develop a class of rules spanning the range between quadratic discriminant
analysis and naive Bayes, through a path of sparse graphical models.
A group lasso penalty is used to introduce shrinkage and encourage
a similar pattern of sparsity across precision matrices. It gives
sparse estimates of interactions and produces interpretable models.
Inspired by the connected-components structure of the estimated precision
matrices, we propose the \textit{community Bayes} model, which partitions
features into several conditional independent communities and splits the classification
problem into separate smaller ones. The community Bayes idea is quite
general and can be applied to non-Gaussian data and likelihood-based
classifiers.
\end{abstract}

\section{Introduction}

In the generic classification problem, the outcome of interest $G$
falls into $K$ unordered classes, which for convenience we denote
by $\{1,2,\ldots,K\}$. Our goal is to build a rule for predicting
the class label of an item based on $p$ measurements of features
$X\in\mathbb{R}^{p}$. The training set consists of the class labels
and features for $n$ items. This is an important practical problem
with applications in many fields. 

Quadratic discriminant analysis (QDA) can be derived as the maximum
likelihood method for normal populations with different means and
different covariance matrices. It is a favored tool when there are
some strong interactions and linear boundaries cannot separate the
classes. However, QDA is poorly posed when the sample size $n_{k}$
is not considerably larger than $p$ for any class $k$, and clearly
ill-posed if $n_{k}\leq p$. Therefore it encourages us to employ
a regularization method \cite{o1986statistical,titterington1985common}.
Although there already exist a number of proposals to regularize linear
discriminant analysis (LDA) \cite{bickel2004some,clemmensen2011sparse,dudoit2002comparison,guo2007regularized,hastie1995penalized,krzanowski1995discriminant,tibshirani2002diagnosis,xu2009modified},
few methods have been developed to regularize QDA. Friedman \cite{friedman1989regularized}
suggests applying a ridge penalty to within-class covariance matrices,
and Price et al. \cite{price2013ridge} propose to add a ridge penalty
and a ridge fusion penalty to the log-likelihood. For both methods,
no elements of the resulting within-class precision matrices will
be zero, leading to dense interaction terms in the discriminant functions
and difficulties in interpretation. Assuming sparse conditions on both unknown means and covariance matrices, Li and Shao \cite{li2013sparse} propose to construct mean and covariance estimators by thresholding. Sun and Zhao \cite{sun2015application} assume block-diagonal structure for covariance matrices and estimate each block by $l_1$ regularization. The blocks are assumed to be of the same size, and are formed based on two sample $t$ statistics. However, both methods can only apply to two-class classification.
Another approach is to assume
conditional independence of the features (naive Bayes). Naive Bayes
classifiers often outperform far more sophisticated alternatives when
the dimension $p$ of the feature space is high , but the conditional
independence assumption may be too rigid in the presence of strong
interactions. 

In this paper we develop a class of rules spanning the range between
QDA and naive Bayes using a \textit{group lasso} penalty \cite{yuan2006model}.
A group lasso penalty is applied to the $(i,j)$th element across
all $K$ within-class precision matrices, which forces the zeros in
the $K$ estimated precision matrices to occur in the same places.
This shared pattern of sparsity results in a sparse estimate of interaction
terms, making the classifier easy to interpret. We refer to this classification
method as \textit{Sparse Quadratic Discriminant Analysis} (SQDA).

The connected components in the resulting precision matrices correspond
exactly to those obtained from simple thresholding rules on the within-class
sample covariance matrices \cite{danaher2013joint}. This suggests
us to partition features into several independent communities before
estimating the parameters \cite{buhlmann2013correlated,hussami2013component,park2007averaged}. 

Therefore, we propose the \textit{community Bayes} model, a simple
idea to identify and make use of conditional independent communities of features.
Furthermore, it is a general approach applicable to non-Gaussian data
and any likelihood-based classifiers. Specifically, the community
Bayes works by partitioning features into several communities, solving
separate classification problems for each community, and combining
the community-wise results into one final prediction. We show that
this approach can improve the accuracy and interpretability of the
corresponding classifier.

The paper is organized as follows. In section \ref{gRDA} we discuss
sparse quadratic discriminant analysis and the sample covariance thresholding
rules. In section \ref{sec:CB} we discuss the community Bayes idea.
Simulated and real data examples appear in Section \ref{sec:examples},
concluding remarks in Section \ref{sec:conclusion} and the proofs
are gathered in the Appendix.

\section{Sparse Quadratic Discriminant Analysis}

\label{gRDA}

Suppose we have training data $(x_{i},g_{i})\in\mathbb{R}^{p}\times K$,
$i=1,2,\ldots,n$, where $x_{i}$'s are observations with measurements
on a set of $p$ features and $g_{i}$'s are class labels. We assume
that the $n_{k}$ observations within the $k$th class are identically
distributed as $N(\mu_{k},\mathbf{\Sigma}_{k})$, and the $n=\sum_{k}n_{k}$
observations are independent. We let $\pi_{k}$ denote the prior for
class $k$. Then the quadratic discriminant function is 
\begin{equation}
\delta_{k}(x)=\frac{1}{2}\log\mbox{det }\mathbf{\Sigma}_{k}^{-1}-\frac{1}{2}(x-\mu_{k})^{T}\mathbf{\Sigma}_{k}^{-1}(x-\mu_{k})+\log\pi_{k}.\label{qdf}
\end{equation}

Let $\mathbf{\Theta}^{(k)}=\mathbf{\Sigma}_{k}^{-1}$ denote the true
precision matrix for class $k$, and $\theta_{ij} = (\theta_{ij}^{(1)}, \ldots, \theta_{ij}^{(K)})$ denote the vector of the $(i,j)$th element across all $K$ precision matrices. We propose to estimate $\mu,\pi,\mathbf{\Theta}$
by maximizing the penalized log likelihood 

\begin{equation}
\sum_{k=1}^{K}\left[\frac{n_{k}}{2}\log\text{det }\mathbf{\Theta}^{(k)}-\frac{1}{2}\sum_{g_{i}=k}(x_{i}-\mu_{k})^{T}\mathbf{\Theta}^{(k)}(x_{i}-\mu_{k})+n_{k}\log\pi_{k}\right] - \frac{\lambda}{2} \sum_{i\neq j}\|\theta_{ij}\|_2.\label{cLikelihood}
\end{equation}
The last term is a \textit{group lasso} penalty, applied to the $(i,j)$th
element across all $K$ precision matrices, which forces a similar
pattern of sparsity across all the precision matrices. Hence, we refer
to this classification method as \textit{Sparse Quadratic Discriminant
Analysis} (SQDA).

Let $\mathbf{\Theta} = (\mathbf{\Theta}^{(1)}, \ldots, \mathbf{\Theta}^{(K)})$. Solving (\ref{cLikelihood}) gives us the estimates
\begin{align}
\hat{\mu}_{k}= & \frac{1}{n_{k}}\sum_{g_{i}=k}x_{i}\\
\hat{\pi}_{k}= & \frac{n_{k}}{n}\\
\hat{\mathbf{\Theta}}= & \mbox{argmax}_{\mathbf{\Theta}^{(k)}\succ0}\sum_{k=1}^{K}n_{k}\left(\log\text{det }\mathbf{\Theta}^{(k)}-\text{tr}(\mathbf{S}^{(k)}\mathbf{\Theta}^{(k)})\right)-\lambda\sum_{i\neq j}\|\theta_{ij}\|_2,\label{grdaEstimate}
\end{align}
where $\mathbf{S}^{(k)}=\frac{1}{n_{k}}\sum_{g_{i}=k}(x_{i}-\hat{\mu}_{k})(x_{i}-\hat{\mu}_{k})^{T}$
is the sample covariance matrix for class $k$.

This model has several important features:
\begin{enumerate}
\item $\lambda$ is a nonnegative tuning parameter controlling the simultaneous
shrinkage of all precision matrices toward diagonal matrices. The
value $\lambda=0$ gives rise to QDA, whereas $\lambda=\infty$ yields
the naive Bayes classifier. Values between these limits represent
degrees of regularization less severe than naive Bayes. Since it is
often the case that even small amounts of regularization can largely
eliminate quite drastic instability, 
smaller values of $\lambda$ (smaller than $\infty$) have the potential
of superior performance when some features have strong partial correlations.
\item The group lasso penalty forces $\hat{\mathbf{\Theta}}^{(1)},\ldots,\hat{\mathbf{\Theta}}^{(K)}$
to share the locations of the nonzero elements, leading to a sparse
estimate of interactions. Therefore, our method produces more interpretable
results.
\item The optimization problem (\ref{grdaEstimate}) can be separated into
independent sub-problems of the same form by simple screening rules
on $\mathbf{S}^{(1)},\ldots,\mathbf{S}^{(K)}$. This leads to a potentially
massive reduction in computational complexity.
\end{enumerate}
The convex optimization problem (\ref{grdaEstimate}) can be quickly
solved by R package JGL \cite{danaher2013joint}. Not for the purpose
of classification as here, Danaher et al. \cite{danaher2013joint}
propose to jointly estimate multiple related Gaussian graphical models
by maximizing the group graphical lasso
\begin{equation}
\mbox{max}_{\mathbf{\Theta}^{(k)}\succ0}\sum_{k=1}^{K}n_{k}\left(\log\mbox{det}\mathbf{\Theta}^{(k)}-\mbox{tr}(\mathbf{S}^{(k)}\mathbf{\Theta}^{(k)})\right)-\lambda_{1}\sum_{i\neq j}\|\theta_{ij}\|_1 -\lambda_{2}\sum_{i\neq j}\|\theta_{ij}\|_2.\label{eq:JGL}
\end{equation}
They use the additional lasso penalty to further encourage sparsity
within $(\theta_{ij}^{(1)},\ldots,\theta_{ij}^{(K)})$. However, for
the purpose of classification, our goal is to identify interaction
terms -- that is, we are interested in whether $(\theta_{ij}^{(1)},\ldots,\theta_{ij}^{(K)})$
is a zero vector instead of whether each element $\theta_{ij}^{(k)}$
is zero. Therefore, we only use the group lasso penalty to estimate
precision matrices.\\

\begin{figure}[h]
\begin{centering}
\includegraphics[scale=0.4]{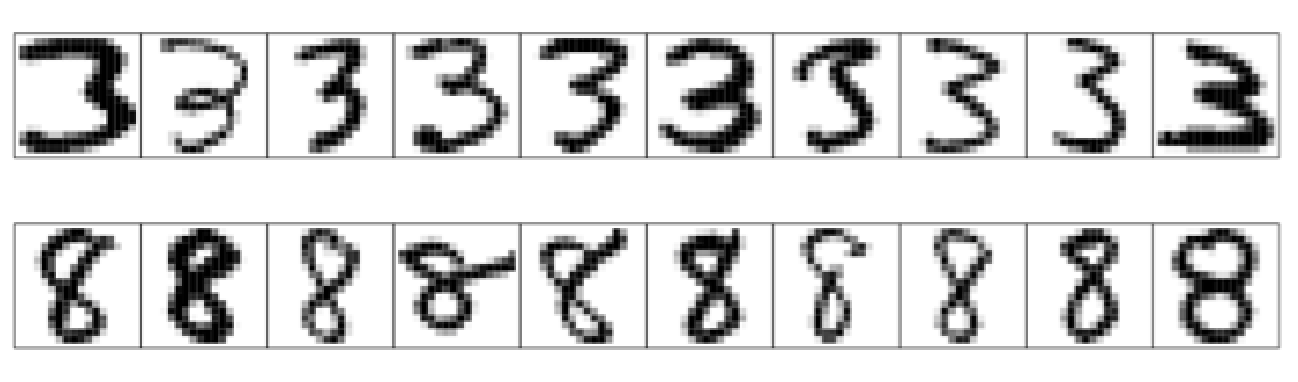}
\par\end{centering}

\caption{\label{fig:3s}\textit{Examples of digitized handwritten 3s and 8s.
Each image is a 8 bit, $16\times16$ gray scale version of the original
binary image.}}
\end{figure}

We illustrate our method with a handwritten digit recognition example. Here we focus on the sub-task
of distinguishing handwritten 3s and 8s. We use the same data as Le
Cun et al. \cite{le1990handwritten}, who normalized binary images
for size and orientation, resulting in 8-bit, $16\times16$ gray scale
images. There are 658 threes and 542 eights in our training set, and
166 test samples for each. Figure \ref{fig:3s} shows a random selection
of 3s and 8s. 

Because of their spatial arrangement the features are highly correlated
and some kind of smoothing or filtering always helps. We filtered
the data by replacing each non-overlapping $2\times2$ pixel block
with its average. This reduces the dimension of the feature space
from 256 to 64.

We compare our method with QDA, naive Bayes and a variant of RDA which we refer to as diagonal regularized discriminant analysis (DRDA). The estimated covariance matrix of DRDA for class $k$ is  
\begin{equation}
\hat{\mathbf{\Sigma}}^{(k)}(\lambda)=(1-\lambda)\mathbf{S}^{(k)}+\lambda\mbox{diag}(\mathbf{S}^{(k)}),\mbox{ with }\lambda\in[0,1].
\end{equation}
The tuning parameters are selected by 5-fold cross validation. Table \ref{tab:zipcode} shows
the misclassification errors of each method and the corresponding standardized tuning parameters $s=P(\hat{\mathbf{\Theta}}(\lambda))/P(\hat{\mathbf{\Theta}}(0))$, where $P(\mathbf{\Theta})=\Sigma_{i\neq j}||\theta_{ij}^{(1)},\ldots,\theta_{ij}^{(K)}||_2$. The results show that SQDA outperforms
DRDA, QDA and naive Bayes. In Figure \ref{fig:zipcode}, we see that
the test error of SQDA decreases dramatically as the model deviates
from naive Bayes and achieves its minimum at a small value of $s$,
while the test error of DRDA keeps decreasing as the model ranges
from naive Bayes to QDA. This is expected since at small values of
$s$, SQDA only includes important interaction terms and shrinks other
noisy terms to 0, but DRDA has all interactions in the model once
deviating from naive Bayes. To better display the estimated precision
matrices, we standardize the estimates to have unit diagonal (the
standardized precision matrix is equal to the partial correlation
matrix up to the sign of off-diagonal entries). Figure \ref{fig:zipcode_precision}
displays the standardized sample precision matrices, and the standardized
SQDA and DRDA estimates. From Figure \ref{fig:zipcode_precision},
we can see that SQDA only includes interactions within a diagonal
band.

\begin{table}[htb!]
\caption{\label{tab:zipcode}\textit{Digit classification results of 3s and
8s. Tuning parameters are selected by 5-fold cross validation.}}

\vspace{2 mm}
\begin{centering}
\scalebox{0.9}{
\begin{tabular}{lcccc}
\hline 
 & {\small{}SQDA} & {\small{}DRDA} & {\small{}QDA} & {\small{}Naive Bayes}\tabularnewline
\hline 
{\small{}Test error} & {\small{}0.042} & {\small{}0.063} & {\small{}0.063} & {\small{}0.160}\tabularnewline
{\small{}Training error} & {\small{}0.024} & {\small{}0.024} & {\small{}0.022} & {\small{}0.119}\tabularnewline
{\small{}CV error} & {\small{}0.028} & {\small{}0.043} & {\small{}--} & {\small{}--}\tabularnewline
{\small{}Standardized tuning parameter} & {\small{}0.016} & {\small{}0.760} & {\small{}--} & {\small{}--}\tabularnewline
\hline 
\end{tabular}
}
\par\end{centering}

\end{table}

\begin{center}
\begin{figure}[htb!]
\begin{centering}
\includegraphics[scale=0.4]{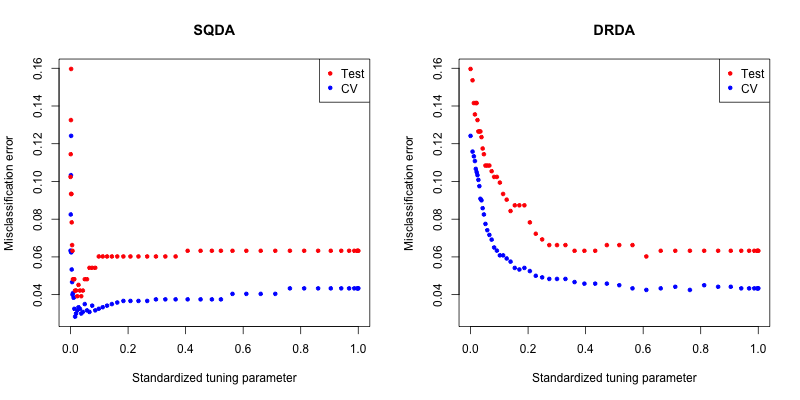}
\par\end{centering}

\caption{\label{fig:zipcode}\textit{The 5-fold cross-validation errors (blue)
and the test errors (red) of SQDA and DRDA on 3s and 8s.}}

\vspace{2 mm}

\begin{centering}
\includegraphics[scale=0.4]{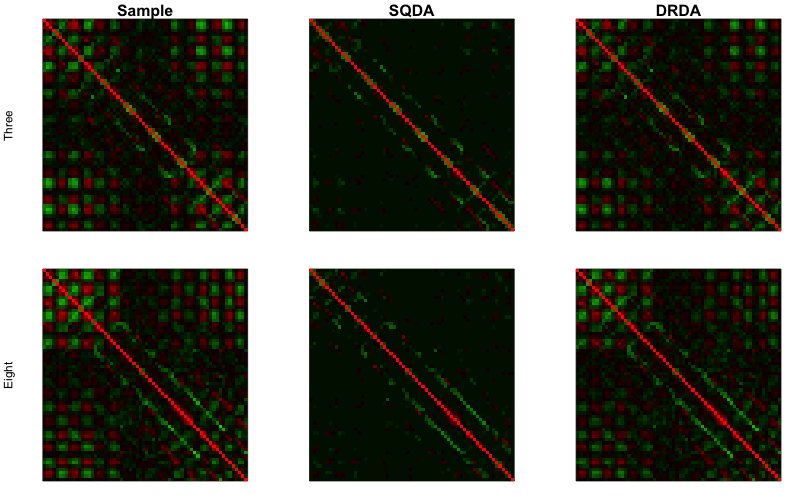}
\par\end{centering}

\caption{\label{fig:zipcode_precision}\textit{Heat maps of the sample precision
matrices and the estimated precision matrices of SQDA and DRDA. Estimates
are standardized to have unit diagonal. The first line corresponds
to the precision matrix of 3s and the second line corresponds to that
of 8s.}}

\end{figure}

\par\end{center}

\subsection{Exact covariance thresholding into connected components}

Mazumder \& Hastie \cite{mazumder2012exact} and Witten et al. \cite{witten2011new}
establish a connection between the graphical lasso and connected components.
Specifically, the connected components in the estimated precision
matrix correspond exactly to those obtained from thresholding the
entries of the sample covariance matrix at $\lambda$. For the solution
to (\ref{grdaEstimate}), we have similar results. By simple thresholding
rules on the sample covariance matrices $\mathbf{S}^{(1)},\ldots,\mathbf{S}^{(K)}$,
the optimization problem (\ref{grdaEstimate}) can be separated into
several optimization sub-problems of the same form, which leads to
huge speed improvements. A more general result for the solution to
(\ref{eq:JGL}) can be found in \cite{danaher2013joint}.

Suppose $\hat{\mathbf{\Theta}}=(\hat{\mathbf{\Theta}}^{(1)},\ldots,\hat{\mathbf{\Theta}}^{(K)})$
is the solution to (\ref{grdaEstimate}) with regularization parameter
$\lambda$. Define 
\begin{equation}
\mathcal{\mathbf{\mathcal{E}}}_{ij}^{(\lambda)}=\begin{cases}
1 & \mbox{if}\ (\hat{\theta}_{ij}^{(1)},\ldots,\hat{\theta}_{ij}^{(K)})\neq\vec{\mathbf{0}},\ i\neq j;\\
0 & \mbox{otherwise}.
\end{cases}\label{eq:concentration1}
\end{equation}
This defines a symmetric graph $\mathcal{G}^{(\lambda)}=(\mathcal{V},\mathcal{\mathbf{\mathcal{E}}}^{(\lambda)})$,
namely the estimated concentration graph defined on the nodes $\mathcal{V}=\{1,\ldots,p\}$
with edges $\mathbf{\mathcal{E}}^{(\lambda)}$. Suppose it admits
a decomposition into $\kappa(\lambda)$ connected components
\begin{equation}
\mathcal{G}^{(\lambda)}=\cup_{l=1}^{\mathbf{\kappa}(\lambda)}\mathcal{G}_{l}^{(\lambda)},\label{eq:concentration2}
\end{equation}
where $\mathcal{\mathcal{G}}_{l}^{(\lambda)}=(\hat{\mathcal{V}}_{l}^{(\lambda)},\mathbf{\mathcal{E}}_{l}^{(\lambda)})$
are the components of the graph $\mathcal{G}^{(\lambda)}$.

Define $\tilde{\mathbf{S}}_{ij}=||(n_{1}\mathbf{S}_{ij}^{(1)},\ldots,n_{K}\mathbf{S}_{ij}^{(K)})||_{2}$.We
can also perform a thresholding on the entries of $\tilde{\mathbf{S}}$
and obtain a graph edge skeleton defined by
\begin{equation}
\mathbf{E}_{ij}^{(\lambda)}=\begin{cases}
1 & \mbox{if}\ \tilde{\mathbf{S}}_{ij}>\lambda,\ i\neq j;\\
0 & \mbox{otherwise}.
\end{cases}\label{eq:threshold1}
\end{equation}
The symmetric matrix $\mathbf{E}^{(\lambda)}$ defines a symmetric
graph $\mathrm{G}^{(\lambda)}=(\mathcal{V},\mathbf{E}^{(\lambda)})$,
which is referred to as the thresholded sample covariance graph. $\mathrm{G}^{(\lambda)}$
also admits a decomposition into connected components
\begin{equation}
\mathrm{G}^{(\lambda)}=\cup_{l=1}^{k(\lambda)}\mathrm{G}_{l}^{(\lambda)},\label{eq:threshold2}
\end{equation}
where $\mathrm{G}_{l}^{(\lambda)}=(\mathcal{V}_{l}^{(\lambda)},\mathbf{E}_{l}^{(\lambda)})$
are the components of the graph $\mathbf{G}^{(\lambda)}$.
\begin{thm}
\label{thm1}For any $\lambda>0$, the components of the estimated
concentration graph $\mathcal{G}^{(\lambda)}$ induce exactly the
same vertex-partition as that of the thresholded sample covariance
graph $\mathrm{G}^{(\lambda)}$. Formally, $\kappa(\lambda)=k(\lambda)$
and there exists a permutation $\pi$ on $\{1,\ldots,k(\lambda)\}$
such that 
\begin{equation}
\hat{\mathcal{V}}_{i}^{(\lambda)}=\mathcal{V}_{\pi(i)}^{(\lambda)},\ \forall i=1,\ldots,k(\lambda).
\label{eq:thm1}
\end{equation}
Furthermore, given $\lambda>\lambda'>0$, the vertex-partition induced
by the components of $\mathcal{G}^{(\lambda)}$ are nested within
that induced by the components of $\mathcal{G}^{(\lambda')}$. That
is, $\kappa(\lambda)\geq\kappa(\lambda')$ and the vertex-partition
$\{\hat{\mathcal{V}}_{l}^{(\lambda)}\}_{1\leq l\leq\kappa(\lambda)}$
forms a finer resolution of $\{\hat{\mathcal{V}}_{l}^{(\lambda')}\}_{1\leq l\leq\kappa(\lambda')}$.\end{thm}
\begin{proof}
The proof of this theorem appears in the Appendix.
\end{proof}
Theorem \ref{thm1} allows us to quickly check the connected components
of the estimated concentration graph by simple screening rules on
$\tilde{\mathbf{S}}$. Notice that for each class $k$, the edge set
$\mathbf{\mathcal{E}}^{(k,\lambda)}$ defined by $\mathbf{\mathcal{E}}_{ij}^{(k,\lambda)}=\mathbf{1}_{\{\hat{\theta}_{ij}^{(k)}\neq0\}}$
is nested in $\mathbf{\mathcal{E}}^{(\lambda)}$. Therefore, the features
can be reordered in such a way that each $\hat{\mathbf{\Theta}}^{(k)}$
is block diagonal
\begin{equation}
\hat{\mathbf{\Theta}}^{(k)}=\left(\begin{array}{cccc}
\hat{\mathbf{\Theta}}_{1}^{(k)} & 0 & \cdots & 0\\
0 & \hat{\mathbf{\Theta}}_{2}^{(k)} & 0 & \cdots\\
\vdots & \vdots & \ddots & \vdots\\
0 & \cdots & 0 & \hat{\mathbf{\Theta}}_{k(\lambda)}^{(k)}
\end{array}\right),
\end{equation}
where the different components represent blocks of indices given by
$\mathcal{V}_{l}^{(\lambda)},\ l=1,\ldots,k(\lambda)$. Then one can
simply solve the optimization problem (\ref{grdaEstimate}) on the
features within each block separately, making problem (\ref{grdaEstimate})
feasible for certain values of $\lambda$ although it may be impossible
to operate on the $p\times p$ variables $\mathbf{\Theta}^{(1)},\ldots,\mathbf{\Theta}^{(K)}$
on a single machine.

\section{Community Bayes}

\label{sec:CB}

The estimated precision matrices of SQDA are often block diagonal
under suitable ordering of the features, which implies that the features
can be partitioned into several communities and these communities
are mutually independent within each class. In this section, we generalize
this idea to non-Gaussian data and other classification models, and
refer to it as the \textit{community Bayes} model. A related work
in the regression setting is \cite{hussami2013component}. The authors
propose to split the lasso problem into smaller ones by estimating
the connected components of the sample covariance matrix. Their approach
involves only one covariance matrix and works specifically for the
lasso.

\subsection{Main idea}

We let $X\in\mathbb{R}^{p}$ denote the feature vector and $G\in\{1,\ldots,K\}$
denote the class variable. Suppose the feature set $\mathcal{V}=\{1,\ldots,p\}$
admits a partition $\mathcal{V}=\cup_{l=1}^{L}\mathcal{V}_{l}$ such
that $X_{\mathcal{\mathcal{V}}_{1}},\ldots,X_{\mathcal{V}_{L}}$ are
mutually independent conditional on $G=k$ for $k=1,\ldots,K$, where
$X_{\mathcal{V}_{l}}$ is a subset of $X$ containing the features
in community $\mathcal{V}_{l}$. Then the posterior probability has
the form
\begin{eqnarray}
\log p(G=k|X) & = & \log p(X|G=k)+\log p(G=k)-\log p(X)\\
 & = & \sum_{l=1}^{L}\log p(X_{\mathcal{V}_{l}}|G=k)+\log p(G=k)-\log p(X)\\
 & = & \sum_{l=1}^{L}\log p(G=k|X_{\mathcal{V}_{l}})+(1-L)\log p(G=k)+C(X)\label{eq:condIndep}
\end{eqnarray}
where $C(X)=\sum_{l=1}^{L}\log p(X_{\mathcal{V}_{l}})-\log p(X)$
only depends on $X$ and serves as a normalization term.

The equation (\ref{eq:condIndep}) implies an interesting result: we can fit the
classification model on each community separately, and combine the
resultant posteriors into the posterior to the original problem by
simply adjusting the intercept and normalizing it. This result has
three important consequences:
\begin{enumerate}
\item The global problem completely separates into $L$ smaller tractable
sub-problems of the same form, making it possible to solve an otherwise
infeasible large-scale problem. Moreover, the modular structure lends
it naturally to parallel computation. That is, one can solve these
sub-problems independently on separate machines. 
\item This idea is quite general and can be applied to any likelihood-based
classifiers, including discriminant analysis, multinomial logistic
regression, generalized additive models, classification trees, etc.
\item When using a classification model with interaction terms, (\ref{eq:condIndep})
doesn't involve interactions across different communities. Therefore,
it has fewer degrees of freedom and thus smaller variance than the
global problem.
\end{enumerate}

To find conditionally independent communities, we use the Spearman's rho, a robust nonparametric rank-based statistics, to directly estimate the unknown correlation matrices, and then apply average, single or complete linkage clustering to the estimated correlation matrices to get the estimated communities. In next section, we will show that this procedure consistently identify conditionally independent communities when the data are from a nonparanormal family. 

The community Bayes algorithm is summarized below.

\begin{algorithm}
\caption{The community Bayes algorithm}
\begin{algorithmic}[1]
\STATE Compute $\hat{\mathbf{R}}^{(k)}$, the estimated correlation matrix
based on the Spearman's rho for each class $k$.
\STATE Perform average, single or complete linkage clustering with similarity
matrix $\tilde{\mathbf{R}}$, where $\tilde{\mathbf{R}}_{ij}=||(n_{1}\hat{\mathbf{R}}_{ij}^{(1)},\ldots,n_{K}\hat{\mathbf{R}}_{ij}^{(K)})||_{2}$,
and cut the dendrogram at level $\tau$ to produce vertex-partition
$\mathcal{V}=\cup_{l=1}^{L}\hat{\mathcal{V}}_{l}$.
\STATE For each community $l=1,\ldots,L$, estimate $\log p(G=k|X_{\hat{\mathcal{V}}_{l}})$
using a classification method of choice.
\STATE Pick $\tau$ and any other tuning parameters by cross-validation.
\end{algorithmic}
\end{algorithm}

\subsection{Community estimation} \label{community}

In this section we address the key part of the community Bayes model:
how to find conditionally independent communities. We propose to derive the communities by applying average, single or complete linkage clustering to the correlation matrices, which are estimated based on nonparametric rank-based statistics. In the case where data are from a nonparanormal family, we prove that given knowledge of $L$, this procedure
consistently identify conditionally independent communities.

We assume $X=(X_{1},\ldots,X_{p})^{T}|G=k$ follows a nonparanormal
distribution $NPN(\mu^{(k)},\mathbf{\Sigma}^{(k)},f^{(k)})$ \cite{liu2009nonparanormal}
and $\mathbf{\Sigma}^{(k)}$ is nonsingular. That is, there exists
a set of univariate strictly increasing transformations $f^{(k)}=\{f_{j}^{(k)}\}_{j=1}^{p}$
such that 
\begin{equation}
Z^{(k)}:=f^{(k)}(X)|G=k\sim N(\mu^{(k)},\mathbf{\Sigma}^{(k)}),
\end{equation}
where $f^{(k)}(X)=(f_{1}^{(k)}(X_{1}),\ldots,f_{p}^{(k)}(X_{p}))^{T}$.
Notice that the transformation functions $f^{(k)}$ can be different
for different classes. To make the model identifiable, $f^{(k)}$
preserves the population mean and standard deviations: $\mathbb{E}(X_{j}|G=k)=E(f_{j}^{(k)}(X_{j})|G=k)=\mu_{j}^{(k)}$,
$\mbox{Var}(X_{j}|G=k)=\mbox{Var}(f_{j}^{(k)}(X_{j})|G=k)={\sigma_{j}^{(k)}}^{2}$.
Liu et al.\cite{liu2009nonparanormal} prove that the nonparanormal
distribution is a Gaussian copula when the transformation functions
are monotone and differentiable.

Let $\mathbf{R}^{(k)}$ denote the correlation matrix of the Gaussian
distribution $Z^{(k)}$, and define 
\begin{equation}
\mathcal{\mathbf{\mathcal{E}}}_{ij}=\begin{cases}
1 & \mbox{if}\ (\mathbf{R}_{ij}^{(1)},\ldots,\mathbf{R}_{ij}^{(K)})\neq\vec{\mathbf{0}},\ i\neq j;\\
0 & \mbox{otherwise}.
\end{cases}\label{eq:corrGraph}
\end{equation}
Suppose the graph $\mathcal{G}=(\mathcal{V},\mathcal{\mathbf{\mathcal{E}}})$
admits a decomposition into $L$ connected components $\mathcal{G}=\cup_{l=1}^{L}\mathcal{G}_{l}.$
Then the vertex-partition $\mathcal{V}=\cup_{l=1}^{L}\mathcal{V}_{l}$
induced by this decomposition gives us exactly the conditionally independent
communities we need. This is because 
\begin{equation}
\mathcal{G}_{l}\mbox{ and }\mathcal{G}_{l'}\mbox{ are disconnected}\Leftrightarrow Z_{\mathcal{V}_{l}}^{(k)}\independent Z_{\mathcal{V}_{l'}}^{(k)},\  \forall k\Leftrightarrow X_{\mathcal{V}_{l}}\independent X_{\mathcal{V}_{l'}}|G=k,\ \forall k.
\end{equation}

Let $(x_{i},g_{i})\in\mathbb{R}^{p}\times K,\ i=1,\ldots,n$ be the
training data where $x_{i}=(x_{i1},\ldots,x_{ip})^{T}$. We estimate
the correlation matrices using Nonparanormal SKEPTIC \cite{liu2012nonparanormal},
which exploits the Spearman's rho and Kendall's tau to directly estimate
the unknown correlation matrices. Since the estimated correlation
matrices based on the Spearman's rho and Kendall's tau have similar
theoretical performance, we only adopt the ones based on the Spearman's
rho here. 

In specific, let $\hat{\rho}_{ij}^{(k)}$ be the Spearman's rho between
features $i$ and $j$ based on $n_{k}$ samples in class $k$, i.e.,
$\{x_{i}|g_{i}=k,i=1,\ldots,n\}$. Then the estimated correlation
matrix for class $k$ is 

\begin{equation}
\hat{\mathbf{R}}_{ij}^{(k)}=\begin{cases}
2\sin(\frac{\pi}{6}\hat{\rho}_{ij}^{(k)}) & i\neq j;\\
1 & i=j.
\end{cases}\label{eq:corrEst}
\end{equation}
Notice that the graph defined by $\mathbf{R}^{(k)}$ has the same
vertex-partition as that defined by $(\mathbf{R}^{(k)})^{-1}$. Inspired
by the exact thresholding result of SQDA, we define $\tilde{\mathbf{R}}_{ij}=||(n_{1}\hat{\mathbf{R}}_{ij}^{(1)},\ldots,n_{K}\hat{\mathbf{R}}_{ij}^{(K)})||_{2}$
and perform exact thresholding on the entries of $\tilde{\mathbf{R}}$
at a certain level $\tau$, where $\tau$ is estimated by cross-validation.
The resultant vertex-partition yields an estimate of the conditionally
independent communities. Furthermore, there is an interesting connection
to hierarchical clustering. Specifically, the vertex-partition induced
by thresholding matrix $\tilde{\mathbf{R}}$ corresponds to the subtrees
from when we apply single linkage agglomerative clustering to $\tilde{\mathbf{R}}$
and then cut the dendrogram at level $\tau$ \cite{tan2013cluster}.
Single linkage clustering tends to produce trailing clusters in which
individual features are merged one at a time. However, Theorem \ref{thm2}
shows that given knowledge of the true number of communities $L$,
application of single, average or complete linkage agglomerative clustering
on $\tilde{\mathbf{R}}$ consistently estimates the vertex-partition
of $\mathcal{G}$.
\begin{thm}
\label{thm2}Assume that $\mathcal{G}$ has $L$ connected components
and $\min_{k}n_{k}\geq\frac{21}{\log p}+2$. Define $\tilde{\mathbf{R}}_{ij}^{0}=||(n_{1}\mathbf{R}_{ij}^{(1)},\ldots,n_{K}\mathbf{R}_{ij}^{(K)})||_{2}$
and let 
\begin{equation}
\min_{i,j\in\mathcal{V}_{l};l=1,\ldots,L}\tilde{\mathbf{R}}_{ij}^{0}\geq16\pi K\sqrt{n_{k}\log p},\mbox{ for }k=1,\ldots,K.\label{eq:consistCond}
\end{equation}
Then the estimated vertex-partition $\mathcal{V}=\cup_{l=1}^{L}\hat{\mathcal{V}}_{l}$
resulting from performing SLC, ALC, or CLC with similarity matrix
$\tilde{\mathbf{R}}$ satisfies $P(\exists l:\hat{\mathcal{V}}_{l}\neq\mathcal{V}_{l})\leq\frac{K}{p^{2}}$.\end{thm}
\begin{proof}
The proof of this theorem appears in the Appendix.
\end{proof}
A sufficient condition for (\ref{eq:consistCond}) is 
\begin{equation}
\min_{i,j\in\mathcal{V}_{l};l=1,\ldots,L}||(\mathbf{R}_{ij}^{(1)},\ldots,\mathbf{R}_{ij}^{(K)})||_{2}\geq16\pi K\sqrt{\frac{\log p}{n_{max}}}\frac{n_{max}}{n_{min}},
\end{equation}
where $n_{min}=\min_{k}n_{k}$ and $n_{max}=\max_{k}n_{k}$. Therefore,
Theorem \ref{thm2} establishes the consistency of identification
of conditionally independent communities by performing hierarchical
clustering using SLC, ALC or CLC, provided that $n_{max}=\Omega(\log p)$,
$n_{max}/n_{min}=O(1)$ as $n_{k},p\rightarrow\infty$, and provided
that no within-community element of $\mathbf{R}^{(k)}$ is too small
in absolute value for all $k$.

\section{Examples}

\label{sec:examples}

\subsection{Sparse quadratic discriminant analysis}

In this section we study the performance of SQDA in several simulated
examples and a real data example. The results show that SQDA achieves
a lower misclassification error and better interpretability compared
to naive bayes, QDA and a variant of RDA. Since RDA proposed by Friedman
\cite{friedman1989regularized} has two tuning parameters, resulting
in an unfair comparison, we use a version of RDA that shrinks $\hat{\mathbf{\Sigma}}^{(k)}$
towards its diagonal: 
\begin{equation}
\hat{\mathbf{\Sigma}}^{(k)}(\lambda)=(1-\lambda)\mathbf{S}^{(k)}+\lambda\mbox{diag}(\mathbf{S}^{(k)}),\mbox{ with }\lambda\in[0,1]
\end{equation}
Note that $\lambda=0$ corresponds to QDA and $\lambda=1$ corresponds
to the naive Bayes classifier. For any $\lambda<1$, the estimated
covariance matrices are dense. We refer to this version of RDA as
diagonal regularized discriminant analysis (DRDA) in the rest of this
paper. 

We report the test misclassification errors and its corresponding
standardized tuning parameters of these four classifiers. Let $P(\mathbf{\Theta})=\sum_{i\neq j}||(\theta_{ij}^{(1)},\ldots,\theta_{ij}^{(K)})||_{2}$.
The standardized tuning parameter is defined as $s=P(\hat{\mathbf{\Theta}}(\lambda))/P(\hat{\mathbf{\Theta}}(0))$,
with $s=0$ corresponding to the naive Bayes classifier and $s=1$
corresponding to QDA.

\subsubsection{Simulated examples}

The data generated in each experiment consists of a training set,
a validation set to tune the parameters, and a test set to evaluate
the performance of our chosen model. Following the notation of \cite{zou2005regularization},
we denote ././. the number of observations in the training, validation
and test sets respectively. For every data set, the tuning parameter
minimizing the validation misclassification error is chosen to compute
the test misclassification error. 

Each experiment was replicated 50 times. In all cases the population
class conditional distributions were normal, the number of classes
was $K=2$, and the prior probability of each class was taken to be
equal. The mean for class $k$ was taken to have Euclidean norm $\sqrt{\mbox{tr}(\mathbf{\Sigma}^{(k)})/p}$
and the two means were orthogonal to each other. We consider three
different models for precision matrices and in all cases the $K$
precision matrices are from the same model:
\begin{itemize}
\item Model 1. Full model: $\theta_{ij}^{(k)}=1$ if $i=j$ and $\theta_{ij}^{(k)}=\rho_{k}$
otherwise.
\item Model 2. Decreasing model: $\theta_{ij}^{(k)}=\rho_{k}^{|i-j|}$.
\item Model 3. Block diagonal model with block size $q$: $\theta_{ij}^{(k)}=1$
if $i=j$, $\theta_{ij}^{(k)}=\rho_{k}$ if $i\neq j$ and $i,j\leq q$
and $\theta_{ij}^{(k)}=0$ otherwise, where $0<q<p$. 
\end{itemize}
Table \ref{tab:dense} and \ref{tab:sparse}, summarizing the results
for each situation, present the average test misclassification errors
over the 50 replications. The quantities in parentheses are the standard
deviations of the respective quantities over the 50 replications.

\subsubsection*{Example 1: Dense interaction terms}

We study the performance of SQDA on Model 1 and Model 2 with $p=8$,
$\rho_{1}=0$, and $\rho_{2}=0.8$. We generated 50/50/10000 observations
for each class. Table \ref{tab:dense} summarizes the results. 

In this example, both situations have full interaction terms and should
favor DRDA. The results show that SQDA and DRDA have similar performance,
and both of them give lower misclassification errors and smaller standard
deviations than QDA or naive Bayes. The model-selection procedure
behaves quite reasonably, choosing large values of the standardized
tuning parameter $s$ for both SQDA and DRDA.

\begin{center}
\begin{table}[h]
\caption{\label{tab:dense}\textit{Misclassification errors and selected tuning
parameters for simulated example 1. The values are averages over 50
replications, with the standard errors in parentheses.}}

\vspace{2 mm}
\begin{centering}
\scalebox{0.9}{
\begin{tabular}{>{\raggedright}m{4cm}>{\raggedright}m{2cm}>{\raggedright}m{2cm}}
\hline 
 & \textbf{\small{}Full} & \textbf{\small{}Decreasing}\tabularnewline
\hline 
\multirow{5}{4cm}{\textbf{\small{}Misclassification error}{\small{}\newline SQDA\newline
DRDA\newline QDA\newline Naive Bayes}} & \multirow{5}{2cm}{{\small{}\newline 0.129(0.021)\newline 0.129(0.020)\newline 0.238(0.028)\newline
0.179(0.026)}} & \multirow{5}{2cm}{{\small{}\newline 0.103(0.011)\newline 0.104(0.013)\newline 0.170(0.031)\newline
0.153(0.032)}}\tabularnewline
 &  & \tabularnewline
 &  & \tabularnewline
 &  & \tabularnewline
 &  & \tabularnewline
\multirow{3}{4cm}{\textbf{\small{}Average standardized tuning}{\small{} }\textbf{\small{}parameter}{\small{}\newline
SQDA\newline DRDA}} & \multirow{3}{2cm}{{\small{}\newline \newline 0.828(0.258)\newline 0.835(0.302)}} & \multirow{3}{2cm}{{\small{}\newline\newline 0.817(0.237)\newline 0.781(0.314)}}\tabularnewline
 &  & \tabularnewline
 &  & \tabularnewline
&  & \tabularnewline
\hline 
\end{tabular}
}
\par\end{centering}

\end{table}

\par\end{center}

\subsubsection*{Example 2: Sparse interaction terms}

We study the performance of SQDA on Model 3 with $\rho_{1}=0$, $\rho_{2}=0.8$
and $q=4$. We performed experiments for $(p,n)=(8,50),(20,200),(40,800),(100,1500)$,
and for each $p$ we generated $n/n/1000$0 observations for each
class. Table \ref{tab:sparse} summarizes the results.

In this example, all situations have sparse interaction terms and
should favor SQDA. Moreover, the sparsity level increases as $p$
increases. As conjectured, SQDA strongly dominates with lower misclassification
errors at all dimensionalities. The standardized tuning parameter
values for DRDA are uniformly larger than those for SQDA. This is
expected because in order to capture the same amount of interactions
DRDA needs to include more noises than SQDA.

\begin{center}
\begin{table}
\caption{\label{tab:sparse}\textit{Misclassification errors and selected tuning
parameters for simulated example 2. The values are averages over 50
replications, with the standard errors in parentheses.}}

\vspace{2 mm}
\begin{centering}
\tabcolsep=0.11cm
\scalebox{0.9}{
\begin{tabular}{>{\raggedright}m{4cm}>{\raggedright}m{2cm}>{\raggedright}m{2cm}>{\raggedright}m{2cm}>{\raggedright}m{2cm}}
\hline 
 & \textbf{\small{}$\boldsymbol{p=8,}$}{\small \par}

\textbf{\small{}$\boldsymbol{n=50}$} & \textbf{\small{}$\boldsymbol{p=20,}$ $\boldsymbol{n=200}$} & \textbf{\small{}$\boldsymbol{p=40,}$ $\boldsymbol{n=800}$} & \textbf{\small{}$\boldsymbol{p=100},$ $\boldsymbol{n=1500}$ }\tabularnewline
\hline 
\multirow{5}{4cm}{\textbf{\small{}Misclassification error}{\small{}\newline SQDA\newline
DRDA\newline QDA\newline Naive Bayes}} & \multirow{5}{2cm}{{\small{}\newline 0.079(0.010)\newline 0.088(0.013)\newline 0.092(0.011)\newline
0.107(0.014)}} & \multirow{5}{2cm}{{\small{}\newline 0.091(0.007)\newline 0.109(0.009)\newline 0.114(0.006)\newline
0.121(0.004)}} & \multirow{5}{2cm}{{\small{}\newline 0.086(0.006)\newline 0.108(0.007)\newline 0.127(0.002)\newline
0.114(0.004)}} & \multirow{5}{2cm}{{\small{}\newline 0.153(0.007)\newline 0.188(0.009)\newline 0.230(0.025)\newline
0.215(0.035)}}\tabularnewline
 &  &  &  & \tabularnewline
 &  &  &  & \tabularnewline
 &  &  &  & \tabularnewline
 &  &  &  & \tabularnewline
\multirow{3}{4cm}{\textbf{\small{}Average standardized tuning}{\small{} }\textbf{\small{}parameter}{\small{}\newline
SQDA\newline DRDA}} & \multirow{3}{2cm}{{\small{}\newline\newline 0.529(0.309)\newline 0.688(0.314)}} & \multirow{3}{2cm}{{\small{}\newline \newline 0.190(0.046)\newline 0.293(0.345)}} & \multirow{3}{2cm}{{\small{}\newline\newline 0.109(0.015)\newline 0.194(0.230)}} & \multirow{3}{2cm}{{\small{}\newline\newline 0.024(0.004)\newline 0.073(0.059)}}\tabularnewline
 &  &  &  & \tabularnewline
  &  &  &  & \tabularnewline
 &  &  &  & \tabularnewline
\hline 
\end{tabular}
}
\par\end{centering}

\end{table}

\par\end{center}

\subsubsection{Real data example: vowel recognition data} \label{sec:vowel}
This data consists of training and test data with 10 predictors and 11 classes. We obtained the data from the benchmark collection maintained by Scott Fahlman at Carnegie Mellon University. The data was contributed by Anthony Robinson \cite{robinson1989dynamic}. The classes correspond to 11 vowel sounds, each contained in 11 different words. Here are the words, preceded by the symbols that represent them:\\

\begin{centering}
\scalebox{0.9}{
\begin{tabular}{ll|ll|ll}
\hline 
 {\small{}Vowel} & {\small{}Word} & {\small{}Vowel} & {\small{}Word} & {\small{}Vowel} & {\small{}Word} 
 \tabularnewline
\hline 
{\small{}i} & {\small{}heed} &  {\small{}a:} & {\small{}hard} &
{\small{}U} & {\small{}hood}
\tabularnewline
{\small{}I} & {\small{}hid} & {\small{}Y} & {\small{}hud} &
{\small{}u:} & {\small{}who'd}
\tabularnewline
{\small{}E} & {\small{}head} & {\small{}O} & {\small{}hod} & 
{\small{}3:} & {\small{}heard}
\tabularnewline
 {\small{}A} & {\small{}had} &  {\small{}C:} & {\small{}hoard}
\tabularnewline
\hline 
\end{tabular}
}
\par\end{centering}
\vspace{2mm}

The word was uttered once by each of the 15 speakers. Four male and four female speakers were used to train the models, and the other four male and three female speakers were used for testing the performance.

This paragraph is technical and describes how the analog speech signals were transformed into a 10-dimensional feature vector. The speech signals were low-pass filtered at 4.7kHz and then digitized to 12 bits with a 10kHz sampling rate. Twelfth-order linear predictive analysis was carried out on six 512-sample Hamming windowed segments from the steady part of the vowel. The reflection coefficients were used to calculate 10 log-area parameters, giving a 10-dimensional input space. Each speaker thus yielded six frames of speech from 11 vowels. This gave 528 frames from the eight speakers used to train the models and 462 frames from the seven speakers used to test the models.

We implement our method on a subset of data, which consists of four vowels "Y", "O", "U" and "u:". These four vowels are selected because their sample precision matrices are approximately sparse, and SQDA usually performs well on such dataset. Figure \ref{fig:vowel_precision} displays the standardized sample precision matrices, and the standardized SQDA and DRDA estimates. In addition to DRDA, QDA and naive Bayes, we also compare our method with three other classifiers, namely the Support Vector Machine (SVM), k-nearest neighborhood (kNN) and Random Forest (RF), in terms of misclassification rate. The SVM, kNN, and Random Forest were implemented by the R packages of "e1071", "class" and "randomForest" with default settings, respectively. The results of these classification procedures are shown in Table \ref{tab:vowel}.

\begin{center}
\begin{figure}[htb!]

\begin{centering}
\includegraphics[scale=0.4]{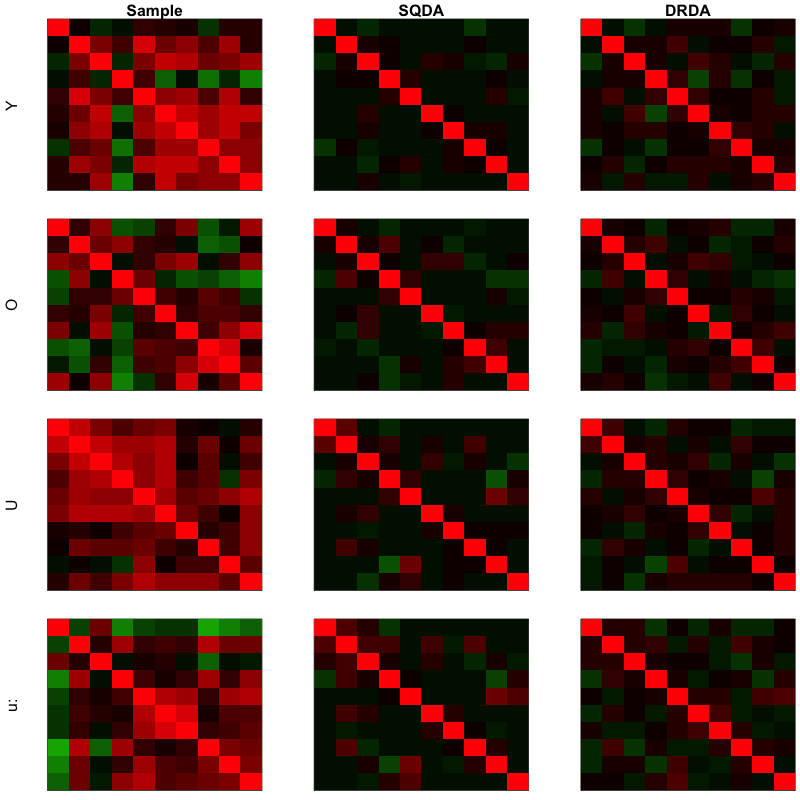}
\par\end{centering}

\caption{\label{fig:vowel_precision}\textit{Heat maps of the sample precision
matrices and the estimated precision matrices of SQDA and DRDA on vowel data. Estimates
are standardized to have unit diagonal.}}

\end{figure}

\par\end{center}

\begin{table}[htb!]
\caption{\label{tab:vowel}\textit{Vowel speech classification results. Tuning parameters are selected by 5-fold cross validation.}}

\vspace{2 mm}
\begin{centering}
\scalebox{0.9}{
\begin{tabular}{lccccccc}
\hline 
 & {\small{}SQDA} & {\small{}DRDA} & {\small{}QDA} & {\small{}Naive Bayes}& {\small{}SVM}& {\small{}kNN}& {\small{}Random Forest}
 \tabularnewline
\hline 
{\small{}Test error} & {\small{}0.172} & {\small{}0.197} & {\small{}0.351} & {\small{}0.304}
& {\small{}0.220} & {\small{}0.274} & {\small{}0.278}
\tabularnewline
{\small{}CV error} & {\small{}0.239} & {\small{}0.260} & {\small{}--} & {\small{}--}& {\small{}--}& {\small{}--}& {\small{}--}\tabularnewline
\hline 
\end{tabular}
}
\par\end{centering}
\end{table}

The SQDA performs significantly better than the other six classifiers. Compared with the second winner DRDA, the SQDA has a relative gain of (19.7\% - 17.2\%)/19.7\% = 12.7\%.

\subsection{Community Bayes}

In this section we study the performance of the community Bayes model
using logistic regression as an example. We refer to the classifier
that combines the community Bayes algorithm with logistic regression
as community logistic regression. We compare the performance of logistic
regression (LR) and community logistic regression (CLR) in several
simulated examples and a real data example, and the results show that
CLR has better accuracy and smaller variance in predictions.

\subsubsection{Simulated examples}

The data generated in each experiment consists of a training set,
a validation set to tune the parameters, and a test set to evaluate
the performance of our chosen model. Each experiment was replicated
50 times. In all cases the distributions of $Z^{(k)}$ were normal,
the number of classes was $K=3$, and the prior probability of each
class was taken to be equal. The mean for class $k$ was taken to
have Euclidean norm $\sqrt{\mbox{tr}(\mathbf{\Sigma}^{(k)})/(p/2)}$
and the three means were orthogonal to each other. We consider three
models for covariance matrices and in all cases the $K$ covariance
matrices are the same:
\begin{itemize}
\item Model 1. Full model: $\mathbf{\Sigma}_{ij}=1$ if $i=j$ and $\mathbf{\Sigma}_{ij}=\rho$
otherwise.
\item Model 2. Decreasing model: $\mathbf{\Sigma}_{ij}=\rho^{|i-j|}$.
\item Model 3. Block diagonal model with $q$ blocks: $\mathbf{\Sigma}=\mbox{diag}(\mathbf{\Sigma}_{1},\ldots,\mathbf{\Sigma}_{q})$.
$\mathbf{\Sigma}_{i}$ is of size $ $$\lfloor\frac{p}{q}\rfloor\times\lfloor\frac{p}{q}\rfloor$
or $(\lfloor\frac{p}{q}\rfloor+1)\times(\lfloor\frac{p}{q}\rfloor+1)$
and is from Model 1.
\end{itemize}
For simplicity, the transformation functions for all dimensions were
the same $f_{1}^{(k)}=\ldots=f_{p}^{(k)}=f^{(k)}$. Define $g^{(k)}=(f^{(k)})^{-1}$.
In addition to the identity transformation, two different transformations
$g^{(k)}$ were employed as in \cite{liu2009nonparanormal}: the symmetric
power transformation and the Gaussian CDF transformation. See \cite{liu2009nonparanormal}
for the definitions. 

We compare the performance of LR and CLR on the above models with
$p=16$, $\rho=0.5$ and $q=2,4,8$. We generated 20/20/10000 observations
for each class and use average linkage clustering to estimate the
communities. Two examples were studied:
\begin{itemize}
\item Example 1: The transformations $g^{(k)}$ for all three classes are
the identity transformation. The population class conditional distributions
are normal with different means and common covariance matrix. The
logits are linear in this example.
\item Example 2: The transformations $g^{(k)}$ for class 1,2 and 3 are
the identity transformation, the symmetric power transformation with
$\alpha=3$, and the Gaussian CDF transformation with $\mu_{g_{0}}=0$
and $\sigma_{g_{0}}=1$ respectively. $\alpha,\mu_{g_{0}}$ and $\sigma_{g_{0}}$
are defined in \cite{liu2009nonparanormal}. The power and CDF transformations
map a univariate normal distribution into a highly skewed and a bi-modal
distribution respectively. The logits are nonlinear in this example.
\end{itemize}
Table \ref{tab:CBsimulation} summarizes the test misclassification
errors of these two methods and the estimated numbers of communities.
CLR performs well in all cases with lower test errors and smaller
standard errors than LR, including the ones where the covariance matrices
don't have a block structure. The community Bayes algorithm introduces
a more significant improvement when the class conditional distributions
are not normal.

\begin{center}
\begin{table}[h]
\caption{\label{tab:CBsimulation}\textit{Misclassification errors and estimated
numbers of communities for the two simulated examples. The values
are averages over 50 replications, with the standard errors in parentheses.}}

\vspace{2 mm}

\begin{centering}
\scalebox{0.9}{
\begin{tabular}{>{\raggedright}m{3.2cm}>{\centering}m{1.5cm}>{\centering}m{1.5cm}>{\centering}m{1.5cm}>{\centering}m{1.5cm}>{\centering}m{1.5cm}}
\hline 
 & \textbf{\small{}Full} & \textbf{\small{}Decreasing} & \multicolumn{3}{c}{\textbf{\small{}Block}}\tabularnewline
 &  &  & {\small{}$q=2$} & {\small{}$q=4$} & {\small{}$q=8$}\tabularnewline
\hline 
\multirow{4}{3.5cm}{\textbf{\small{}Example 1}{\small{}\newline CLR}\inputencoding{latin1}{{\small{}
test error}}\inputencoding{latin9}{\small{}\newline LR test error\newline
Number of communities}} & \multirow{4}{1.5cm}{{\small{}\newline 0.253(0.041)\newline 0.274(0.040)\newline 5.220(4.292)}} & \multirow{4}{1.5cm}{{\small{}\newline 0.221(0.032)\newline 0.286(0.036)\newline 6.040(4.262)}} & \multirow{4}{1.5cm}{{\small{}\newline 0.227(0.033)\newline 0.271(0.039)\newline 4.700(3.882)}} & \multirow{4}{1.5cm}{{\small{}\newline 0.203(0.035)\newline 0.278(0.041)\newline 5.320(3.395)}} & \multirow{4}{1.5cm}{{\small{}\newline 0.262(0.030)\newline 0.325(0.036)\newline 7.780(3.935)}}\tabularnewline
 &  &  &  &  & \tabularnewline
 &  &  &  &  & \tabularnewline
 &  &  &  &  & \tabularnewline
  &  &  &  &  & \tabularnewline
\multirow{3}{3.5cm}{\textbf{\small{}Example 2}{\small{}\newline CLR}\inputencoding{latin1}{
test error}\inputencoding{latin9}{\small{}\newline LR test error\newline
Number of communites}} & \multirow{3}{1.5cm}{{\small{}\newline 0.221(0.049)\newline 0.255(0.061)\newline 7.480(4.253)}} & \multirow{3}{1.5cm}{{\small{}\newline 0.227(0.040)\newline 0.282(0.062)\newline 7.320(3.977)}} & \multirow{3}{1.5cm}{{\small{}\newline 0.208(0.041)\newline 0.263(0.062)\newline 6.940(4.533)}} & \multirow{3}{1.5cm}{{\small{}\newline 0.203(0.036)\newline 0.266(0.063)\newline 6.220(3.627)}} & \multirow{3}{1.5cm}{{\small{}\newline 0.239(0.031)\newline 0.309(0.057)\newline 9.160(3.782)}}\tabularnewline
 &  &  &  &  & \tabularnewline
  &  &  &  &  & \tabularnewline
 &  &  &  &  & \tabularnewline
  &  &  &  &  & \tabularnewline
\hline 
\end{tabular}
}
\par\end{centering}

\end{table}

\par\end{center}

\subsubsection{Real data example: email spam}

The data for this example consists of information from 4601 email
messages, in a study to screen email for ``spam''. The true outcome
\texttt{\small{}email} or\texttt{\small{} spam} is available, along
with the relative frequencies of 57 of the most commonly occurring
words and punctuation marks in the email message. Since most of the spam predictors have a very long-tailed
distribution, we log-transformed each variable (actually $\log(x+0.1)$)
before fitting the models. 

We compare CLR with LR and four other commonly used classifiers, namely SVM, kNN, Random Forest and Boosting. Same as in section \ref{sec:vowel}, the SVM, kNN, Random Forest and Boosting were implemented by the R packages of "e1071", "class", "randomForest" and "ada" with default settings, respectively. The range of the number of communities
for CLR were fixed to be between 1 and 20, and the optimal number of communities were selected
by 5-fold cross validation. We randomly chose
1000 samples from this data, and split them into equally sized training
and test sets. We repeated the random sampling and splitting 20 times. The mean misclassification percentage of each method is listed in Table \ref{tab:spam} with standard error in parenthesis.

\begin{table}[htb!]
\caption{\label{tab:spam}\textit{Email spam classification results. The values are test misclassification errors averaged over 20 replications, with standard errors in parentheses.}}

\vspace{2 mm}
\begin{centering}
\scalebox{0.9}{
\begin{tabular}{cccccc}
\hline 
 {\small{}CLR} & {\small{}LR} & {\small{}SVM}& {\small{}kNN}& {\small{}Random Forest} & {\small{}Boosting}
 \tabularnewline
\hline 
 {\small{}0.068(0.019)} & {\small{}0.087(0.026)} & {\small{}0.070(0.011)} & {\small{}0.106(0.017)}
& {\small{}0.071(0.010)} & {\small{}0.075(0.012)} \tabularnewline
\hline 
\end{tabular}
}
\par\end{centering}
\end{table}

\begin{figure}[h]
\begin{centering}
\includegraphics[scale=0.5]{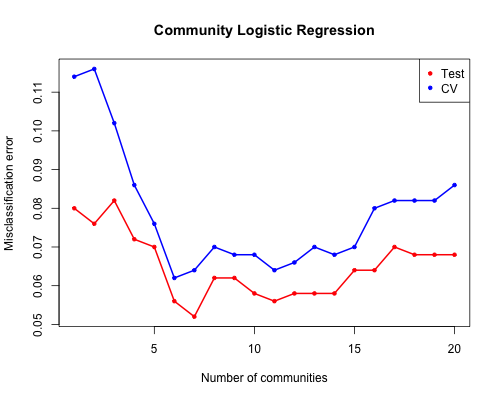}
\par\end{centering}

\caption{\label{fig:spam}\textit{The 5-fold cross-validation errors (blue)
and the test errors (red) of CLR on the spam data. }}

\end{figure}

The mean test error rate for LR is 8.7\%. By comparison, CLR has a mean test
error rate of 6.8\%, yielding a 21.8\% improvement. Figure \ref{fig:spam}
shows the test error and the cross-validation error of CLR over the
range of the number of communities for one replication. It corresponds to LR when the
number of communities is 1. The estimated number of communities in this replication is
6, and these communities are listed below. Most features in community
1 are negatively correlated with spam, while most features in community
2 are positively correlated. 
\begin{itemize}
\item \texttt{\small{}hp, hpl, george, lab, labs, telnet, technology,
direct, original, pm, cs, re, edu, conference, 650, 857, 415, 85, 1999,
ch;, ch(, ch{[},\\ CAPAVE, CAPMAX, CAPTOT.}{\small \par}
\item \texttt{\small{}over, remove, internet, order, free,
money, credit, business, email, mail, receive, will, people, report,
address, addresses, make, all, \\ our, you, your, 000, ch!, ch\$.}{\small \par}
\item \texttt{\small{}font, ch\#.}{\small \par}
\item \texttt{\small{}data, project.}{\small \par}
\item \texttt{\small{}parts, meeting, table.}{\small \par}
\item \texttt{\small{}3d.}{\small \par}
\end{itemize}

\section{Conclusions}

\label{sec:conclusion}In this paper we have proposed sparse quadratic
discriminant analysis, a classifier ranging between QDA and naive
Bayes through a path of sparse graphical models. By allowing interaction
terms into the model, this classifier relaxes the strict and often
unreasonable assumption of naive Bayes. Moreover, the resulting estimates
of interactions are sparse and easier to interpret compared to other
existing classifiers that regularize QDA. 

Motivated by the connection between the estimated precision matrices
and single linkage clustering with the sample covariance matrices,
we present the community Bayes model, a simple procedure applicable
to non-Gaussian data and any likelihood-based classifiers. By exploiting
the block-diagonal structure of the estimated correlation matrices,
we reduce the problem into several sub-problems that can be solved
independently. Simulated and real data examples show that the community
Bayes model can improve accuracy and reduce variance in predictions.

\section*{A Proofs}

\subsection*{A.1 Proof of Theorem \ref{thm1}}
\begin{proof}
The standard KKT conditions \cite{boyd2009convex} of the optimization problem (\ref{grdaEstimate}) give:
\begin{equation}
-n_{k}\left(\left(\hat{\mathbf{\Theta}}^{(k)}\right)^{-1}+\mathbf{S}^{(k)}\right)+\lambda\mathbf{\Gamma}^{(k)}=0,\mbox{ for }k=1,\ldots,K
\end{equation}
where $\mathbf{\Gamma}^{(k)}$ is a $p\times p$ matrix with diagonal
elements being zero: 
\[
\mathbf{\Gamma}^{(k)}=\left(\begin{array}{cccc}
0 & \gamma_{12}^{(k)} & \cdots & \gamma_{1p}^{(k)}\\
\gamma_{21}^{(k)} & 0 & \cdots & \gamma_{2p}^{(k)}\\
\vdots & \vdots & \ddots & \vdots\\
\gamma_{p1}^{(k)} & \gamma_{p2}^{(k)} & \cdots & 0
\end{array}\right).
\]
Denote $\gamma_{ij}=(\gamma_{ij}^{(1)},\dots,\gamma_{ij}^{(K)})$
and $\hat{\theta}_{ij}=(\hat{\theta}_{ij}^{(1)},\ldots,\hat{\theta}_{ij}^{(K)})$.
The subgradient $\gamma_{ij}\in\partial||\hat{\theta}_{ij}||_{2}$
for $i\neq j$, and 
\[
\partial||\theta_{ij}||_{2}=\begin{cases}
B_{2}(1) & \mbox{if }\theta_{ij}=0;\\
\frac{\theta_{ij}}{||\theta_{ij}||_{2}} & \mbox{otherwise.}
\end{cases}
\]

Let $\hat{\mathbf{W}}^{(k)}=\left(\hat{\mathbf{\Theta}}^{(k)}\right)^{-1}.$
Then
\begin{eqnarray}
\left(n_{1}(\hat{\mathbf{W}}_{ij}^{(1)}-\mathbf{S}_{ij}^{(1)}),\ldots,n_{K}(\hat{\mathbf{W}}_{ij}^{(K)}-\mathbf{S}_{ij}^{(K)})\right) & = & \lambda\frac{\hat{\theta}_{ij}}{||\hat{\theta}_{ij}||_{2}},\ \mbox{if }i\neq j\mbox{ and }\hat{\theta}_{ij}\neq0;\\
\left\lVert\left(n_{1}(\hat{\mathbf{W}}_{ij}^{(1)}-\mathbf{S}_{ij}^{(1)}),\ldots,n_{K}(\hat{\mathbf{W}}_{ij}^{(K)}-\mathbf{S}_{ij}^{(K)})\right)\right\rVert_{2} & \leq & \lambda,\ \mbox{if }i\neq j\mbox{ and }\hat{\theta}_{ij}=0;\label{eq:kkt2}\\
\left(\hat{\mathbf{W}}_{ii}^{(1)},\ldots,\hat{\mathbf{W}}_{ii}^{(K)}\right) & = & \left(\mathbf{S}_{ii}^{(1)},\ldots,\mathbf{S}_{ii}^{(K)}\right).
\end{eqnarray}

There exists an ordering of the vertices $\{1,\ldots,p\}$ such that
the edge-matrix of the thresholded covariance graph is block-diagonal.
For notational convenience, we will assume that the matrix is already
in this order:
\begin{equation}
\mathbf{E}^{(\lambda)}=\left(\begin{array}{ccc}
\mathbf{E}_{1}^{(\lambda)} & \cdots & 0\\
\vdots & \ddots & \vdots\\
0 & \cdots & \mathbf{E}_{k(\lambda)}^{(\lambda)}
\end{array}\right),
\end{equation}
where $\mathbf{E}_{l}^{(\lambda)}$ represents the block of indices
given by $\mathcal{V}_{l}^{(\lambda)},l=1,\ldots,k(\lambda)$.

We will construct $\hat{\mathbf{W}}^{(1)},\ldots,\hat{\mathbf{W}}^{(K)}$
having the same structure as $\mathbf{E}^{(\lambda)}$ which is a
solution to the optimization problem (\ref{grdaEstimate}). Note that if $\hat{\mathbf{W}}^{(k)}$
is block diagonal then so is its inverse. Let $\hat{\mathbf{W}}^{(k)}$
and its inverse $\hat{\mathbf{\Theta}}^{(k)}$ be given by:
\begin{equation}
\hat{\mathbf{W}}{}^{(k)}=\left(\begin{array}{ccc}
\hat{\mathbf{W}}_{1}^{(k)} & \cdots & 0\\
\vdots & \ddots & \vdots\\
0 & \cdots & \hat{\mathbf{W}}_{k(\lambda)}^{(k)}
\end{array}\right),\ \hat{\mathbf{\Theta}}^{(k)}=\left(\begin{array}{ccc}
\hat{\mathbf{\Theta}}_{1}^{(k)} & \cdots & 0\\
\vdots & \ddots & \vdots\\
0 & \cdots & \hat{\mathbf{\Theta}}_{k(\lambda)}^{(k)}
\end{array}\right),\mbox{ for all } k.\label{eq:blockD}
\end{equation}
Define $\hat{\mathbf{W}}_{l}^{(k)}$ or equivalently $\hat{\mathbf{\Theta}}_{l}^{(k)}=\left(\hat{\mathbf{W}}_{l}^{(k)}\right)^{-1}$
via the following sub-problems
\begin{equation}
\hat{\mathbf{\Theta}}_l =\mbox{argmin}_{\mathbf{\Theta}_{l}^{(k)} \succ 0}\left\{ -\sum_{k=1}^{K}n_{k}\left(\mbox{logdet }\mathbf{\Theta}_{l}^{(k)}-\mbox{tr}(\mathbf{S}_{l}^{(k)}\mathbf{\Theta}_{l}^{(k)})\right)+\lambda\sum_{\substack{i,j\in\mathcal{V}_{l}^{(\lambda)} \\ i\neq j}}\|\theta_{ij}\|_{2}\right\} 
\end{equation}
for $l=1,\ldots,k(\lambda)$, where $\hat{\mathbf{\Theta}}_l = (\hat{\mathbf{\Theta}}_l^{(1)}, \ldots, \hat{\mathbf{\Theta}}_l^{(K)})$, and $\mathbf{S}_{l}^{(k)}$ is a sub-block
of $\mathbf{S}^{(k)}$ with row/column indices from $\mathcal{V}_{l}^{(\lambda)}\times\mathcal{V}_{l}^{(\lambda)}$.
Thus for every $l$, $(\hat{\mathbf{\Theta}}_{l}^{(1)},\ldots,\hat{\mathbf{\Theta}}_{l}^{(K)})$
satisfies the KKT conditions corresponding to the $l^{\mbox{th}}$
block of the $p\times p$ dimensional problem. 

By construction of the thresholded sample covariance graph, if $i\in\mathcal{V}_{l}^{(\lambda)}$and
$j\in\mathcal{V}_{l'}^{(\lambda)}$ with $l\neq l'$, then $\tilde{\mathbf{S}}_{ij}\leq\lambda$.
Hence, the choice $\hat{\mathbf{\Theta}}_{ij}^{(k)}=\hat{\mathbf{W}}_{ij}^{(k)}=0,\ k=1,\ldots,K$
satisfies the KKT conditions (\ref{eq:kkt2}) for all the off-diagonal
entries in the block-matrix (\ref{eq:blockD}). Hence $(\hat{\mathbf{\Theta}}^{(1)},\ldots,\hat{\mathbf{\Theta}}^{(K)})$
solves the optimization problem (\ref{grdaEstimate}). This shows that the vertex-partition
$\{\hat{\mathcal{V}}_{l}^{(\lambda)}\}_{1\leq l\leq\kappa(\lambda)}$
obtained from the estimated precision graph is a finer resolution
of that obtained from the thresholded covariance graph, i.e., for
every $l\in\{1,\ldots,k(\lambda)\}$, there is a $l'\in\{1,\ldots,\kappa(\lambda)\}$
such that $\hat{\mathcal{V}}_{l'}^{(\lambda)}\subset\mathcal{V}_{l}^{(\lambda)}$.
In particular $k(\lambda)\leq\kappa(\lambda)$.

Conversely, if $(\hat{\mathbf{\Theta}}^{(1)},\ldots,\hat{\mathbf{\Theta}}^{(K)})$
admits the decomposition as in the statement of the theorem, then
it follows from (\ref{eq:kkt2}) that: for $i\in\hat{\mathcal{V}}_{l}^{(\lambda)}$and
$j\in\mathcal{\hat{V}}_{l'}^{(\lambda)}$ with $l\neq l'$, we have
$||\tilde{\mathbf{S}}_{ij}||_{2}\leq\lambda$ since $\hat{\mathbf{W}}_{ij}^{(k)}=0$
for $k=1,\ldots,K$. Thus the vertex-partition induced by the connected
components of the thresholded covariance graph is a finer resolution
of that induced by the connected components of the estimated precision
graph. In particular this implies that $k(\lambda)\geq\kappa(\lambda)$.

Combining the above two we conclude $k(\lambda)=\kappa(\lambda)$
and also the equality (\ref{eq:thm1}). Since the labeling of the connected components
is not unique, there is a permutation $\pi$ in the theorem. 

Note that the connected components of the thresholded sample covariance
graph $\mbox{G}^{(\lambda)}$ are nested within the connected components
of $\mbox{G}^{(\lambda')}$. In particular, the vertex-partition of
the thresholded sample covariance graph at $\lambda$ is also nested
within that of the thresholded sample covariance graph at $\lambda'$.
We have already proved that the vertex-partition induced by the connected
components of the estimated precision graph and the thresholded sample
covariance graph are equal. Using this result, we conclude that the
vertex-partition induced by the components of the estimated precision
graph at $\lambda$, given by $\{\hat{\mathcal{V}}_{l}^{(\lambda)}\}_{1\leq l\leq\kappa(\lambda)}$
is contained inside the vertex-partition induced by the components
of the estimated precision graph at $\lambda'$, given by $\{\hat{\mathcal{V}}_{l}^{(\lambda')}\}_{1\leq l\leq\kappa(\lambda')}$.
\end{proof}

\subsection*{A.2 Proof of Theorem \ref{thm2}}
\begin{proof}
Let $a=\min_{i,j\in\mathcal{V}_{l};l=1,\ldots,L}\tilde{\mathbf{R}}_{ij}^{0}$.
First note that $\{\max_{ij}\left|\tilde{\mathbf{R}}_{ij}-\tilde{\mathbf{R}}_{ij}^{0}\right|<\frac{a}{2}\}\Rightarrow\{\hat{\mathcal{V}}_{l}=\mathcal{V}_{l},\forall l\}$
\cite{tan2013cluster}. Then 
\begin{eqnarray*}
P\left(\exists l,\hat{\mathcal{V}}_{l}\neq\mathcal{V}_{l}\right) & \leq & P\left(\max_{ij}\left|\tilde{\mathbf{R}}_{ij}-\tilde{\mathbf{R}}_{ij}^{0}\right|\geq\frac{a}{2}\right)\\
 & \leq & P\left(\sum_{k=1}^{K}\max_{ij}n_{k}\left||\hat{\mathbf{R}}_{ij}^{(k)}|-|\mathbf{R}_{ij}^{(k)}|\right|\geq\frac{a}{2}\right)\\
 & \leq & \sum_{k=1}^{K}P\left(\max_{ij}n_{k}\left||\hat{\mathbf{R}}_{ij}^{(k)}|-|\mathbf{R}_{ij}^{(k)}|\right|\geq\frac{a}{2K}\right)\\
 & \leq & \sum_{k=1}^{K}P\left(\max_{ij}\left|\hat{\mathbf{R}}_{ij}^{(k)}-\mathbf{R}_{ij}^{(k)}\right|\geq\frac{a}{2n_{k}K}\right)
\end{eqnarray*}

The second inequality is because $\left|\sqrt{a_{1}^{2}+\ldots+a_{m}^{2}}-\sqrt{b_{1}^{2}+\ldots+b_{m}^{2}}\right|\leq\left|a_{1}-b_{1}\right|+\ldots+\left|a_{m}-b_{m}\right|$.

Since $a\geq16\pi K\sqrt{n_{k}\log p},$ we have $P(\max_{ij}\left|\hat{\mathbf{R}}_{ij}^{(k)}-\mathbf{R}_{ij}^{(k)}\right|\geq\frac{a}{2n_{k}K})\leq\frac{1}{p^{2}}$
by Theorem 4.1 of \cite{liu2012nonparanormal}. Therefore $P(\exists l,\hat{\mathcal{V}}_{l}\neq\mathcal{V}_{l})\leq\frac{K}{p^{2}}$.\end{proof}

\bibliographystyle{plain}
\bibliography{mybib}

\end{document}